\RequirePackage{xcolor}
\documentclass[journal,twoside,web]{ieeecolor}

\usepackage{generic}
\usepackage{cite}
\usepackage{amsmath,amssymb,amsfonts}
\usepackage{graphicx}
\usepackage{textcomp}
\usepackage{hyperref}
\hypersetup{hidelinks=true}

\usepackage{booktabs}
\usepackage{multirow}
\usepackage{pgfplots}
\usepackage{soul}
\usepackage{url}
\usepackage{algorithm}
\usepackage{algorithmic}
\usepackage{float}
\usepackage{caption}
\usepackage{subcaption}
\usepackage{diagbox}

\def\BibTeX{{\rm B\kern-.05em{\sc i\kern-.025em b}\kern-.08em
    T\kern-.1667em\lower.7ex\hbox{E}\kern-.125emX}}
\DeclareMathOperator*{\argmin}{argmin}

\newcommand{\mat}[1]{\ensuremath{\operatorname{\boldsymbol{\MakeUppercase{#1}}}}} 
\newcommand{\norm}[1]{\left \lVert #1\right \rVert}    

\newcommand{\R}{\ensuremath{\mathbb{R}}}    
\newcommand{\inRe}[1]{\in\R^{#1}} 
\newcommand{\tens}[1]
{\boldsymbol{\mathcal{#1}}} 
\newcommand{\vect}[1]{\boldsymbol{#1}} 
\newcommand{\fact}[2]{\mat{#1}^{(#2)}}
\newcommand{\set}[1]{\mathcal{#1}}

\renewcommand{\O}[1]{\mathcal{O}\left( #1 \right)}
\newcommand{\out}{\otimes}
\newcommand{\khatrao}{\ensuremath{\odot}}
\newcommand{\hadamard}{\ensuremath{\circledast}}
\newcommand{\dotprod}[2]{\ensuremath{\left\langle #1, #2 \right\rangle}}

\makeatletter
\newcommand\bighadamard{\mathop{\mathpalette\b@gCircledast\relax}}
\newcommand\b@gCircledast[2]{%
\vcenter{\hbox{\m@th
\scalebox{\ifx#1\displaystyle 2.1\else1.2\fi}{$#1\circledast$}%
}}\vcenter{\hbox{\rule{0pt}{16pt}}}%
}
\makeatother

\pgfplotsset{compat=1.9}
\usepgfplotslibrary{external}
\usetikzlibrary{external}
\begin{document}

\title{Adapting Tensor Kernel Machines to Enable Efficient 
Transfer Learning for Seizure Detection\\
\author{S. J. S. de Rooij, \IEEEmembership{Student Member, IEEE},  and B. Hunyadi, \IEEEmembership{Senior Member, IEEE}
\thanks{
  S.J.S. de Rooij and B. Hunyadi are with the Signal Processing Systems group in the Microelectronics department at the Faculty of Electrical Engineering, Mathematics and Computer Science (EEMCS), Delft University of Technology, The Netherlands.
  \\
  \indent S.J.S. de Rooij, and thereby this work, is supported by the TU Delft AI initiative.
}}}

\maketitle

\begin{abstract}
Transfer learning aims to optimize performance in a target task by learning from a related source problem. In this work, we propose an efficient transfer learning method using a tensor kernel machine. Our method takes inspiration from the adaptive SVM and hence transfers `knowledge' from the source to the `adapted' model via regularization. The main advantage of using tensor kernel machines is that they leverage low-rank tensor networks to learn a compact non-linear model in the primal domain. This allows for a more efficient adaptation without adding more parameters to the model.
To demonstrate the effectiveness of our approach, we apply the adaptive tensor kernel machine (Adapt-TKM) to seizure detection on behind-the-ear EEG. By personalizing patient-independent models with a small amount of patient-specific data, the patient-adapted model (which utilizes the Adapt-TKM), achieves better performance compared to the patient-independent and fully patient-specific models. Notably, it is able to do so while requiring around 100 times fewer parameters than the adaptive SVM model, leading to a correspondingly faster inference speed. This makes the Adapt-TKM especially useful for resource-constrained wearable devices. 
\end{abstract}

\begin{IEEEkeywords}
transfer learning, tensors, kernel machines, seizure detection
\end{IEEEkeywords}

\section{Introduction}
\IEEEPARstart{I}{nspired} by how humans use prior information to learn to do new tasks, transfer learning aims to improve performance in a \textit{target} task by leveraging information from a related, or \textit{source}, problem \cite{zhuang2021ComprehensiveSurveyTransfer, hosna2022TransferLearningFriendly, pan2010SurveyTransferLearning}. This way, the performance of machine learning models can be improved for target tasks for which there is little to no `information' (i.e. \textit{(supervised) data}) available. 

Such situations arise especially in healthcare applications, where both data collection and labelling, which needs to be done by medical experts, are costly \cite{balakrishnan2012ScalablePersonalizationLongTerm}. For this reason, generalized (or \textit{patient-independent}) models are easier to implement, even though there is significant inter-subject variability with respect to disease processes \cite{ goetz2018PersonalizedMedicineMotivation}. Transfer learning can improve the performance of these generalized models through \textit{personalization}, using only a small amount of patient-specific data to adapt the patient-independent model \cite{wan2021ReviewTransferLearning, decooman2020PersonalizingHeartRateBased}.  

In this work, we focus especially on the problem of electroencephalogram (EEG)-based epileptic seizure detection. Epilepsy is a highly heterogeneous disease, where the seizure patterns observed on the EEG vary substantially between patients, depending on the subtype and origin of seizures in the brain \cite{scheffer2017ILAEClassificationEpilepsies}. Transfer learning (TL) has been applied previously to improve seizure detection and (the related) epilepsy detection \cite{cui2023TransferLearningBased}.

Most of the work using TL for seizure and epilepsy detection considers the semi-supervised or \textit{transductive} TL case. These models consider there to be no labels available for the target dataset and only use \textit{supervision} for the source data. Thus, they typically aim to minimize the distance between the distribution of the source and target data. 

In this manner, in \cite{yang2014TransductiveDomainAdaptive}, the large-margin-projected transductive SVM (LMPROJ) is used to improve epilepsy detection. The LMPROJ model minimizes the maximum mean discrepancy (MMD) between source and target data during training. Inspired by this work, the authors of \cite{jiang2017SeizureClassificationEEG} developed a semi-supervised learning algorithm using the TSK fuzzy system for model interpretability.
Instead of minimizing the distance during the training step, the authors of \cite{jiang2019TransferComponentAnalysis} use transfer component analysis to learn a feature latent subspace where distances between patients are minimized. The classifier is then trained in this feature subspace.  

A concern for the implementability of these models may be that the data from the source patients needs to be available for training the TL model, which could raise privacy concerns. Therefore, in \cite{zhao2023SourceFreeDomainAdaptation}, the authors develop a privacy-preserving seizure subtype classification model which uses source-free domain adaptation. This way, data from the source patients do not need to be saved. 

The downside of these transductive TL (or semi-supervised) models is that they are only able to minimize the distance between the marginal distributions of the data. However, there can also be conditional distribution differences in epileptic EEG signals, i.e. differences in the distributions conditioned on the labels \cite{cui2023TransferLearningBased}. Methods to account for these differences require labels in the target domain (i.e. \textit{inductive} TL).  

In \cite{page2016WearableSeizureDetection}, they implemented a max-pooling CNN (MPCNN) to perform end-to-end learning for seizure detection. They found that fine-tuning the patient-independent model with patient-specific data outperformed both the patient-independent (PI) and the patient-specific (PS) model. Their approach is similar to our previous work, where we used fine-tuning on tensor kernel machines to update a PI model with PS data \cite{derooij2024EfficientPatientFineTuned}. However, in that case, the fine-tuned model did not outperform the PS model.

Alternatively, in  \cite{decooman2020PersonalizingHeartRateBased} the authors use the adaptive SVM model to personalize a heart-rate based seizure detection algorithm. This adaptive SVM model \cite{yang2007AdaptingSVMClassifiers} uses the distance to the source model (in this case, the PI model) as a regularizer to train a new model for the target. This way, an \textit{adapted} model is learned that inherits the support vectors from the \textit{source} and adds new ones from the \textit{target} data. The downside of this is that it increases the model size, which can be especially problematic if the PI model is already large (in the case of a large PI training set). 

Moreover, recently, there has been an increase in interest in the development of efficient machine learning models \cite{schwartz2020GreenAI}. Both from a sustainability perspective, as well as for application in edge devices \cite{mao2024GreenEdgeAI} such as wearables for health monitoring. To this end, low-rank tensor networks have been proposed as a valuable asset to enable this `Green AI' \cite{memmel2024PositionTensorNetworks}. These tensor networks allow for a parameter-efficient expression of the models' parameters by using low-rank approximations. 

Therefore, in this paper, we propose an \textit{efficient} transfer learning method using tensor kernel machines \cite{wesel2021LargeScaleLearningFourier, klus2019TensorbasedAlgorithmsImage} based on the adaptive SVM model \cite{yang2007AdaptingSVMClassifiers}. 
To show the effectiveness of the method, we apply it to the detection of epileptic seizures on behind-the-ear EEG \cite{chatzichristos2023SeizeIT1}. Wearable behind-the-ear EEG devices are being developed to provide continuous, noninvasive monitoring for people with epilepsy \cite{macea2023InhospitalHomebasedLongterm, lehnen2025RealTimeSeizureDetection, bhagubai2025SeizeIT2WearableDataset}. They aim to improve quality of life by enabling better tracking of seizures and providing a warning of seizure onset \cite{komal2024SystematicReviewLiterature, hadady2023RealworldUserExperience, vandecasteele2020VisualSeizureAnnotation}. 

The adaptability of our model makes it possible to improve detection performance via personalization, even when limited patient-specific data is available.
Furthermore, the model's efficiency makes it especially useful for these resource-constrained devices \cite{komal2024SystematicReviewLiterature}. It may allow for on-device inference as well as adaptation, which could reduce (or even eliminate) the need for data broadcasting. 

The rest of this paper is structured as follows. First, in Section \ref{sec:kernel_machine_and_tensors} background information is provided on kernel machines and tensor networks. Then, in the following section (\ref{sec:adapt_TKM}), we introduce the adaptive tensor kernel machine (Adapt-TKM). After which, we describe the seizure detection pipeline (Section \ref{sec:seizure_detection}). Results for the Adapt-TKM are provided in Section \ref{sec:experiments_results}. These results are discussed in Section \ref{sec:discussion}, and some suggestions are made for future work. Finally, Section \ref{sec:conclusion} concludes this paper.

\section{Kernel Machines and Tensors} \label{sec:kernel_machine_and_tensors}

\subsection{Notation and Preliminaries}\label{subsec:notation}

Throughout the paper, the following notation conventions are used. Vector and matrices are denoted by boldface lowercase ($\vect{a}\inRe{I}$) and boldfase uppercase ($\mat{A}\inRe{I\times J}$) letters, respectively. 
Tensors, i.e. \textit{multidimensional arrays} \cite{kolda2009TensorDecompositionsApplications}, are denoted by boldface calligraphic letters  ($\tens{A}\inRe{I_1 \times I_2 \times \cdots \times I_D}$). 
The symbols $\out$, $\khatrao$ and $\hadamard$ are used to denote the tensor outer product, the (column-wise) Khatri-Rao product and the Hadamard product, respectively.

The Frobenius inner product between two tensors is defined as, \mbox{$\langle \tens{A}, \tens{B} \rangle_\mathrm{F}:=\mathrm{vec}(\tens{A})^T\mathrm{vec}(\tens{B})$}, where \mbox{$\mathrm{vec}(\tens{A})_i = a_{i_1 i_2 \ldots i_D}$} is the vectorization of \mbox{$\tens{A} \inRe{I_1 \times \cdots \times I_D}$} to $\vect{a}\inRe{I}$ where $I=\prod_{d=1}^D I_d$. When it comes to vectorization, we use the little-endian ordering convention, as it is most frequently used in tensor network literature \cite{cichocki2016TensorNetworksDimensionalitya}.

A $D$-dimensional tensor $\tens{A} \inRe{I_1 \times I_2 \times \cdots \times I_D}$ is considered to be \textit{rank-one} if it can be written as the outer product of $D$ vectors \cite{kolda2009TensorDecompositionsApplications}, $$\tens{A} = \vect{a}^{(1)} \out \vect{a}^{(2)} \out \cdots \out \vect{a}^{(D)} = \bigotimes_{d=1}^D \vect{a}^{(d)}.$$

Because the number of elements in a tensor grows exponentially with the dimension, low-rank tensor networks are typically used to represent the tensor with a reduced number of parameters \cite{cichocki2016TensorNetworksDimensionalitya}. 
The two tensor networks most often used for this purpose are the tensor-train decomposition (TT) \cite{oseledets2011TensorTrainDecomposition} and the canonical polyadic decomposition (CPD) \cite{kruskal1977ThreewayArraysRank, hitchcock1927ExpressionTensorPolyadic}. 

A rank-$R$ CPD decomposes a tensor $\tens{A}\inRe{I_1 \times I_2 \times \cdots \times I_D}$ into a sum of $R$ rank-one tensors,

\begin{equation}
\tens{A} :=  \sum_{r=1}^R \gamma_r \bigotimes_{d=1}^D \vect{a}_r^{(d)},\label{eq:cpd} 
\end{equation}
where the vectors $\vect{a}_r^{(d)}$ are normalized to unit norm and the scaling is absorbed by $\gamma_r$. 

The CPD can also be expressed as the Khatri-Rao product of \textit{factor matrices}, $$\text{vec}(\tens{A}):= \left(\mat{A}^{(D)} \khatrao \mat{A}^{(D-1)} \khatrao\cdots \khatrao \mat{A}^{(1)} \right) \vect{\gamma} \ .$$
Here, $\vect{\gamma} \inRe{R}$ and the columns of the factor matrices contain the vectors of the rank-one components, i.e. \mbox{$\mat{A}^{(d)} := [ \vect{a}_1^{(d)} \ \vect{a}_2^{(d)} \ \cdots \ \vect{a}_R^{(d)} ] \inRe{I_d \times R}$}.

\subsection{Supervised Learning and Kernel Machines}
In supervised learning the aim is to find a function $f(\cdot)$ that maps the input data $\vect{x}\in \set{X}$ to its output $y\in \set{Y}$, given a dataset of $N$ observations of input-output pairs $\{\vect{x}_n, y_n\}_{n=1}^{N}$. 
Kernel machines perform nonlinear regression ($y\in \R$) or classification ($y \in \{-1,1\}$) by mapping the input data into a higher dimensional feature space via a feature map \mbox{$\Phi(\cdot): \set{X} \rightarrow \set{H}$} \cite{scholkopf2002LearningKernelsSupport}. In this feature space, they search for a linear relationship between the features. 
Kernel machines thus have the following model form\footnote{
Note that we omit a bias term here. This is done to simplify notation and because a bias term is not necessarily needed when the data is properly scaled \cite{chang2011LIBSVMLibrarySupport}. However, a bias term can always be added if needed. 
}:
\begin{equation}
    f(x) = \langle \Phi(\vect{x}), \vect{w} \rangle
\end{equation}

The weights of the model, $\vect{w}$, can be obtained by minimizing the (regularized) empirical risk,
\begin{equation}
    \vect{w} = \argmin_{\vect{w}} \left[ \frac{1}{N} \sum_{n=1}^N \ell \left(\dotprod{\Phi(\vect{x}_n)}{\vect{w}}, y_n \right) + \lambda \norm{\vect{w}}_p^2 \right]
\end{equation}
where $\ell(\cdot): \set{X} \times \set{Y} \rightarrow \R^+$ is a convex loss function. Different loss functions lead to the primal formulation of different kernel machines. The perhaps most well-known kernel machine, the SVM \cite{cortes1995SupportvectorNetworks}, uses the hinge loss function in its primal. Kernel ridge regression (or LS-SVMs) \cite{suykens1999LeastSquaresSupport}, on the other hand, use the squared loss function. 

Due to the high dimensionality induced by the feature maps, most kernel machines are optimized in the dual (which can be derived using Lagrange multipliers) where the feature maps only appear in the model via their inner products. These inner products can be computed with kernel functions that operate in a Hilbert space: $\kappa(\vect{x}_i, \vect{x}_j) = \langle \Phi(\vect{x}_i), \Phi(\vect{x}_j) \rangle$ \cite{scholkopf2002LearningKernelsSupport}. Thus, eliminating the need for explicit high-dimensional feature maps.

The downside of this dual formulation is that the size of the kernel matrix ($\mat{K}_{i,j} = \kappa(\vect{x}_i, \vect{x}_j)$), which contains the output of the kernel function for each combination of input samples, scales exponentially which the number of input samples ($\O{N^2}$). This means that to solve the dual and find the $\O{N}$ dual coefficients ($\hat{\alpha}_n$) an optimization procedure is needed that is at least $\O{N^2}$ \cite{chang2011LIBSVMLibrarySupport}. 
Furthermore, for model inference \eqref{eq:dual_kernel_machine_inference}, it is necessary to store and use (most of) the training set ($\O{ND}$) as the \textit{support vectors}. All of this could be prohibitively expensive for large enough $N$.
\begin{equation}
f(x) =\sum_{n=1}^{N} \hat{\alpha}_n y_n \kappa(\vect{x}, \vect{x}_n)
\label{eq:dual_kernel_machine_inference}
\end{equation}

Therefore, the tensor kernel machines we consider in this paper\footnote{There also exist support tensor machines \cite{guo2016SupportTensorMachines}, which take tensorial data as input. Note that these are different from the tensor kernel machines we discuss in this paper.} \cite{wesel2021LargeScaleLearningFourier, klus2019TensorbasedAlgorithmsImage} are optimized in the primal, and the challenges that arise from the high dimensionality of the feature space are tackled using low-rank tensor networks. 
Thus, we consider feature maps that are the outer product of $D$ local feature maps, $\phi^{(d)}(\cdot): \R \rightarrow \R^{M_d}$,  
\begin{align}
    \mat{\Phi}(\vect{x}) = \bigotimes_{d=1}^D \vect{\phi}^{(d)}(x^{(d)}) ,
    \label{eq:rank1_features}
\end{align}
which results in a rank-1 tensor $\Phi(\vect{x}) \inRe{M_1\times M_2 \times \cdots \times M_D}$ for input data $\vect{x}\inRe{D}$.
These tensor feature maps can approximate a whole range of functions by using pure power polynomials or Fourier features as the local feature maps \cite{wesel2024QuantizedFourierPolynomial, solin2020HilbertSpaceMethods}. One such function is the frequently used RBF kernel (Section \ref{subsec:feat_map_hyper_param_tune}) \cite{wesel2021LargeScaleLearningFourier, rasmussen2005GaussianProcessesMachine}.

Using these tensor feature maps, the weights of the kernel machine can be expressed as a tensor, $\tens{W}\inRe{M_1\times M_1 \times \cdots \times M_D}$, leading to the primal of the tensor kernel machine (TKM) with rank-1 feature maps \cite{wesel2021LargeScaleLearningFourier, klus2019TensorbasedAlgorithmsImage},
\begin{equation}
\begin{aligned}
    &\min_{\tens{W}} \frac{1}{N} \sum_{n=1}^N \ell \left(\dotprod{\mat{\Phi}(\vect{x}_n)}{\tens{W}}_\textrm{F}, y_n \right) + \lambda \norm{\tens{W}}_\textrm{F}^2 \\
    &\text{s.t.  }
     \mat{\Phi}(\vect{x}) = \bigotimes_{d=1}^D \vect{\phi}^{(d)}(x^{(d)})\ .
\end{aligned} \label{eq:primal_TKM}
\end{equation}

This weight tensor scales exponentially with $D$ (the input data size). Therefore, in TKMs the weight tensor is typically approximated using a low-rank tensor network, such as a CPD or tensor-train.
These tensor networks allow for linear \cite{oseledets2011TensorTrainDecomposition, kolda2009TensorDecompositionsApplications}, or even log-linear \cite{khoromskij2011OdlogNQuanticsApproximation, wesel2024QuantizedFourierPolynomial},  scaling with the input data size.

Throughout, this paper we will use the CPD for the weights, since this decomposition has fewer parameters than the tensor-train decomposition and is thus more efficient to learn.
Furthermore, we will use the squared error as loss function, this leads to tensor kernel ridge regression (TKRR) \cite{wesel2021LargeScaleLearningFourier},
\begin{equation}
\begin{aligned}
     &\min_{\tens{W}} \ \frac{1}{N} \sum_{n=1}^N  \left(\dotprod{\mat{\Phi}(\vect{x}_n)}{\tens{W}}_\text{F}- y_n \right)^2 + \lambda \norm{\tens{W}}_\textrm{F}^2 \\
     &\text{s.t.  }
     \mat{\Phi}(\vect{x}) = \bigotimes_{d=1}^D \vect{\phi}^{(d)}(x^{(d)}), \ \tens{W} = \sum_{r=1}^R \gamma_r \bigotimes_{d=1}^D \vect{w}_r^{(d)} 
\end{aligned} \label{eq:primal_CP_KRR}
\end{equation}

This can be efficiently solved using an block coordinate descent procedure, where the factor matrices of the CPD weight tensor are updated alternatingly. The update of this block coordinate descent is done by solving the following least-squares problem, 
\begin{align}
    \begin{split}
            {\min_{\tiny{\text{vec}(\mat{W}^{(d)})}} \ \frac{1}{N}\sum_{n=1}^N\left(\left\langle \vect{g}^{(d)}(\vect{x}_n),\textrm{vec}(\mat{W}^{(d)})\right\rangle_{\mathrm{F}} -y_n\right)^2} \\ {+ \lambda\left\langle\textrm{vec}\left(\mat{W}^{(d)^T}\mat{W}^{(d)}\right), \textrm{vec}\left(\mat{H}^{(d)}\right)\right\rangle_{\mathrm{F}}},
    \end{split}
    \label{eq:als_step_TKRR}
\end{align}
where $\text{vec}\left(\fact{W}{d}\right)$ is the vectorization of the $d$-th factor matrix and, 
\begin{align}
&\vect{g^{(d)}}(\vect{x}) = \text{vec}\left( \vect{\phi}^{(d)}(\vect{x}) \left( \bighadamard_{i=1, i\neq d}^D \vect{\phi}^{(i)}(\vect{x})^{\mathrm{T}} \mat{W}^{(i)} \right)^T \right) \label{eq:def_gd_step_TKRR} \\
&\mat{H}^{(d)} = \left(\bighadamard_{i=1, i\neq d}^D \mat{W}^{(i)^{T}} \mat{W}^{(i)}\right)^T.  \label{eq:def_Hd_step_TKRR}
\end{align}

\subsection{Transfer Learning and the Adaptive SVM} \label{subsec:TL_and_adapt_SVM}
In transfer learning knowledge from a related domain (\textit{source domain}) is used to improve the performance in a target domain \cite{zhuang2021ComprehensiveSurveyTransfer}. There are many different categories of transfer learning \cite{niu2020DecadeSurveyTransfer}, in this paper we focus on the case of inductive transfer learning. 
Inductive refers to the availability of labels in both the source and target domain.  

In \cite{yang2007AdaptingSVMClassifiers} the authors developed an inductive transfer learning method to adapt an existing SVM model (the \textit{source}) to new data (\textit{target}) called Adapt-SVM. In the Adapt-SVM model, the regularization of the weights for the model trained on target data, $\norm{\vect{w}}_{\mathrm{F}}^2$, is replaced by the squared distance between the source and target (or \textit{adapted}) weights, $\norm{\vect{w}^s - \vect{w}}_{\mathrm{F}}^2$. This way, the source model $f^s(x)$ can be considered as a `prior' for the adapted model $f^a(x)$. The resulting model will have additional \textit{support vectors} from the target set, 
$$f^a(x) = f^s(x) + \sum_{n=1}^{N^t} \hat{\alpha}_n y_n \kappa(x, x_n).$$

Thus, the model size of the adapted model is even larger than the source model ($\O{(N^s+N^t)D}$ instead of $\O{N^s D}$). 
This Adapt-SVM method has been used for various applications, including video concept detection \cite{yang2007CrossDomainVideoConcept} and seizure detection using heart rate measurements \cite{decooman2020PersonalizingHeartRateBased}.
\section{Adaptive Tensor Kernel Machine}
\label{sec:adapt_TKM}

In this paper, we propose an adaptive tensor kernel machine (Adapt-TKM) that takes inspiration from the Adapt-SVM model \cite{yang2007AdaptingSVMClassifiers}. Thus, we change the regularization of the TKM \eqref{eq:primal_TKM} to the squared Frobenius distance between the source and the \textit{adapted} weights, 
\begin{align}
        \min_{\tiny{\tens{W}}} \ \frac{1}{N^{t}}\sum_{n=1}^{N^t} \ell \left(\left\langle \mat{\Phi}(\vect{x}^t_n),\tens{W}\right\rangle_{\mathrm{F}}, y_n^t\right) + \mu \norm{\tens{W} - \tens{W}^{s}}_F^2,
    \label{eq:objective_adapt_TKRR}    
\end{align}
where $\tens{W}^{s}$ are the weights of the source model, $\tens{W}$ the weights of the adapted model, and with $\{\vect{x}_n^t, y_n^t\}_{n=1}^{N^t}$ as the target dataset. 
The regularization parameter $\mu$ controls the distance between the source and adapted model weights. 

This objective \eqref{eq:objective_adapt_TKRR} can be solved using block coordinate descent if $\tens{W}$ is a low-rank tensor-network. To derive the update step, we first expand the norm,

\begin{align}
\begin{split}
    &\norm{\tens{W} - \tens{W}^{s}}_{\mathrm{F}}^2 = \langle \tens{W} - \tens{W}^s, \tens{W} - \tens{W}^s \rangle_{\mathrm{F}} \\ & \quad = \langle\tens{W}, \tens{W}\rangle_{\mathrm{F}} - 2 \langle\tens{W}, \tens{W}^s\rangle_{\mathrm{F}} + \langle\tens{W}^s, \tens{W}^s\rangle_{\mathrm{F}}.
\end{split}
\end{align}

From this expansion, we see that $\langle\tens{W}, \tens{W}\rangle_{\mathrm{F}}$ is the same as the regularization term from the `regular' TKRR \eqref{eq:primal_CP_KRR}, and 
$\langle\tens{W}^s, \tens{W}^s\rangle_{\mathrm{F}}$ is constant with respect to $\tens{W}$ and therefore has no influence on the optimal value of $\tens{W}$. 
Therefore, the only `new' term in the optimization procedure is $\langle\tens{W}, \tens{W}^s\rangle_{\mathrm{F}}$. 

If both $\tens{W}^s$ and $\tens{W}$ are CPD tensors, we can separate out the $d$-th factor matrix of $\tens{W}$ for the update step. First, to simplify notation we define $\tens{S}=\tens{W}^s$ with CPD-rank $P$. The dot product $\langle\tens{W}, \tens{S}\rangle_{\mathrm{F}}$ can then be rewritten as follows,

\begin{align}
\begin{split}
&\langle\tens{W}, \tens{S}\rangle_{\mathrm{F}}  =\left\langle \sum_{r=1}^R \bigotimes_{d=1}^D \vect{w}_r^{(d)},
 \sum_{p=1}^P \bigotimes_{d=1}^D \vect{s}_r^{(d)} \right\rangle_{\mathrm{F}} \\
& =\sum_{m_1=1}^{\hat{M}} \cdots \sum_{m_D=1}^{\hat{M}} \sum_{r=1}^R \sum_{p=1}^P w_{m_1 r}^{(1)} s_{m_1, p}^{(1)} \cdots w_{m_D, r}^{(D)} s_{m_D, p}^{(D)} \\ 
& =\operatorname{vec}\left(\vect{W}^{(d)}\right)^{\mathrm{T}} \operatorname{vec}\left({\mat{S}^{(d)}}\left( \bighadamard_{i=1, i\neq d}^D \mat{W}^{(i)^{\mathrm{T}}} \mat{S}^{(i)} \right)^{\!\mathrm{T}\,}\right) \\
& =\left\langle\operatorname{vec}\left(\mat{W}^{(d)} \right), \operatorname{vec}\left(\vect{Q}^{(d)}\right)\right\rangle_{\mathrm{F}}.
\end{split} \label{eq:dot_Wt_Ws}
\end{align}

Combining this into the block coordinate descent step of the TKRR model \eqref{eq:als_step_adapt_tkrr} we obtain the following update step for the Adapt-TKRR model,
\begin{align} 
    \begin{split}
    \min_{\text{vec}(\mat{W}^{(d)})} \quad & \frac{1}{N} \sum_{n=1}^N\left(\left\langle \vect{g}^{(d)}(\vect{x}_n),\textrm{vec}(\mat{W}^{(d)})\right\rangle_{\mathrm{F}} -y_n\right)^2 \\
    &+ \mu\left\langle\textrm{vec}\left(\mat{W}^{(d)^\mathrm{T}}\mat{W}^{(d)}\right), \textrm{vec}\left(\mat{H}^{(d)}\right)\right\rangle_{\mathrm{F}}\\ &-2 \mu \left\langle\operatorname{vec}\left(\mat{W}^{(d)} \right), \operatorname{vec}\left(\vect{Q}^{(d)}\right)\right\rangle_{\mathrm{F}},
    \end{split}
    \label{eq:als_step_adapt_tkrr}
\end{align}
where $\vect{g}^{(d)}(\vect{x})$, $\mat{H}^{(d)}$ and $\mat{Q}^{(d)}$ are defined as in \eqref{eq:def_gd_step_TKRR}, \eqref{eq:def_Hd_step_TKRR} and  \eqref{eq:dot_Wt_Ws}.
Algorithm \ref{alg:adapt-TKM} shows the full Adapt-TKRR algorithm. 

\begin{algorithm}[t]  
\caption{Adapt-TKM}  
\label{alg:adapt-TKM}
\begin{algorithmic}[0]  
\STATE \textbf{Input:} Training data: $\{\vect{x}_n, y_n \}_{n=1}^N$. \\
Regularization parameter, $\mu$, maximum number of iterations, $n_{\text{max}}$, feature map, $\mat{\Phi}$, weights of the source model, $\{\mat{S}^{(d)}\}_{d=1}^D$.
\end{algorithmic}
\begin{algorithmic}[1]
\STATE \textbf{Initialize:} Randomly initialize of the CPD weights, $\{\mat{W}^{(d)}_0\}_{d=1}^D$, or set it equal to the source weights $\{\mat{W}^{(d)}_0\}_{d=1}^D = \{\mat{S}^{(d)}\}_{d=1}^D$
\REPEAT
    \FOR{$d=1:D$}
        \STATE Compute $\forall n \ \vect{g}^{(d)}(\vect{x}_n)$ as defined in \eqref{eq:def_gd_step_TKRR}. 
        \STATE Set the regularization matrices $\mat{H}^{(d)}$ and $\mat{Q}^{(d)}$ as in \eqref{eq:def_Hd_step_TKRR} and  \eqref{eq:dot_Wt_Ws}.
        \STATE Set $\fact{W}{d}$ to solution of \eqref{eq:als_step_adapt_tkrr}.
        \STATE Normalize $\fact{W}{d}$ and set the scaling factors to $\vect{\gamma}$.
        \STATE Update $n_{\text{iter}}+=1$
    \ENDFOR
\UNTIL{$n_{\text{iter}}=n_\text{max}$}

\RETURN $[\vect{\gamma}; \fact{W}{1}, \fact{W}{2}, \ldots, \fact{W}{D} ]$
\end{algorithmic}
\end{algorithm}

\section{Seizure Detection} 
\label{sec:seizure_detection}
\begin{figure*}[t]
    \centering
    \includegraphics[width=0.8\textwidth]{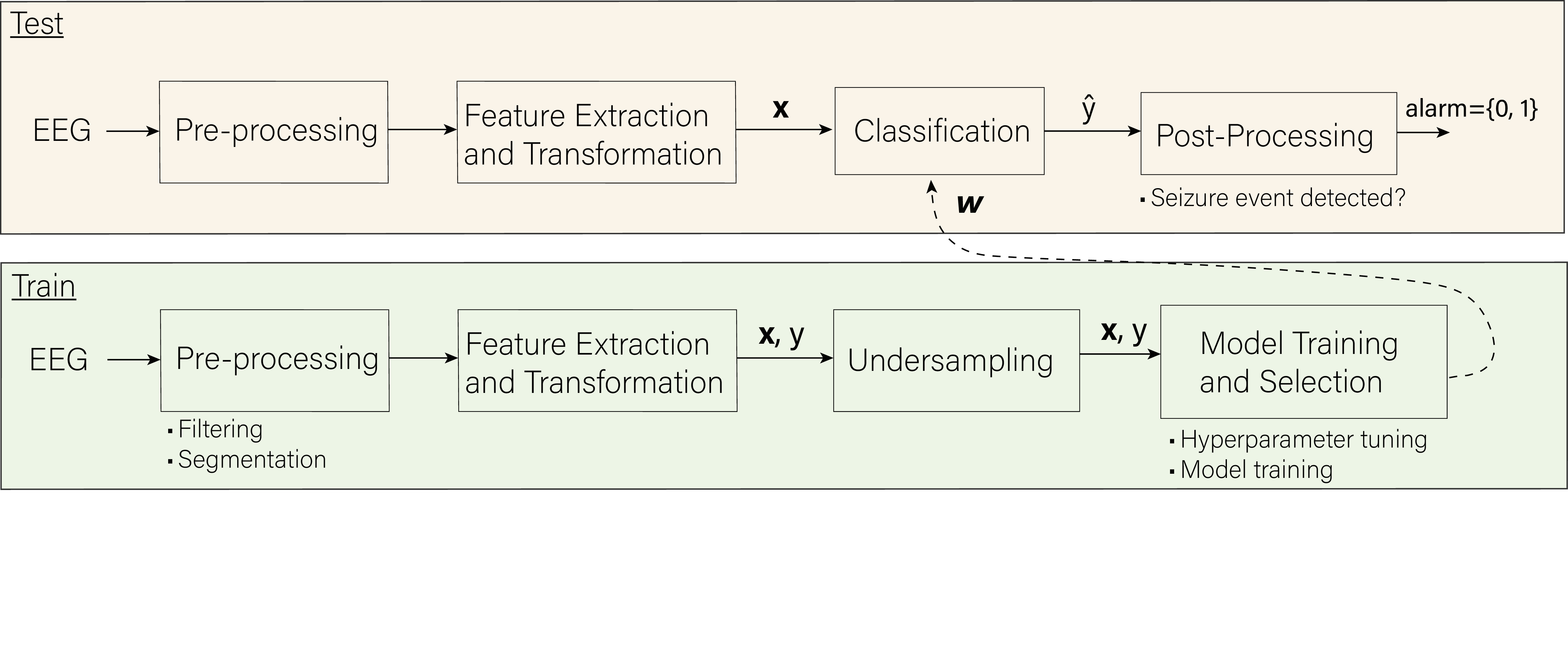}
    \vspace{-5em}
    \caption{Seizure detection pipeline.}
    \label{fig:pipeline_seizure_detection}
\end{figure*}

In this section we describe the seizure detection pipeline. A schematic of this pipeline can be found in Figure \ref{fig:pipeline_seizure_detection}.
We aimed to keep the pipeline close to that of  \cite{vandecasteele2020VisualSeizureAnnotation}, a previous work on seizure detection with behind-the-ear EEG using an SVM classifier.

\subsection{Preprocessing}
The EEG recordings are first band-pass filtered between $1$-$25$Hz. Then, the EEG data is divided into $2$s segments. For testing $50\%$ overlap was used between the segments. However, for the training data different to reduce the data imbalance different amounts of overlap were used for seizure and non-seizure segments. For seizure data $90\%$ overlap was used and for the non-seizure data no overlap was used. Segments with an RMS amplitude less than $13\mu V$ or larger than $150\mu V$ were removed, as they contained mainly background EEG or high-amplitude artifacts \cite{vandecasteele2020VisualSeizureAnnotation}.

\begin{table}[h]
    \centering
    \caption{Extracted features \cite{vandecasteele2020VisualSeizureAnnotation}}
    \setlength{\tabcolsep}{3pt}
    \begin{tabular}{p{2.5cm} p{5cm}}
    \toprule
    \multirow{5}{*}{Time domain} & 1-3. Number of zero crossings, maxima and minima\\
    & 4. Skewness\\
    & 5. Kurtosis\\
    & 6. RMS amplitude\\
    \midrule
    \multirow{3}{*}{\parbox{2.5cm}{Frequency domain}} & 7. Total power\\
    & 8. Peak frequency\\
    & \parbox{5cm}{9-18. Mean and normalized power in frequency bands: $\delta$ (1-3 Hz), $\theta$ (4-8 Hz), $\alpha$ (9-13 Hz), $\beta$ (14-20 Hz), HF (40-80 Hz).}\\
    \midrule
    \multirow{2}{*}{Entropy} & 19. Spectral entropy\\
    & 20. Sample entropy\\
    & 21. Shannon entropy\\
    \bottomrule
    \end{tabular}
    \label{tab:features}
\end{table}

\subsection{Feature Extraction and Transformation}
To reduce the data size and facilitate classification, we extract features for each segment. These features are a combination of frequently used time domain, frequency domain and non-linear entropy-based features \cite{vandecasteele2020VisualSeizureAnnotation}. The features are extracted per EEG channel, thus we end up with a total of $21 \times n_{ch}$ features. 

Similar to what was done in \cite{vandecasteele2020VisualSeizureAnnotation} we use a power transformation, in our case the Yeo-Johnson transformation \cite{yeo2000NewFamilyPower}, to improve the `normality' of certain features: [5-7, 9-13, 15,16,18,21].
Lastly, all features were scaled to lie in the range of $[-0.5, 0.5]$.

\subsection{Class Weighting and Undersampling} \label{subsec:class_weighting_undersamp}
Even though we use more overlap for the seizure segments than the non-seizure segments, the data is still highly imbalanced. Learning with imbalanced data is a challenging topic as it `destroys' the i.i.d assumption \cite{haixiang2017LearningClassimbalancedData}. Thus, we use a combination of undersampling and class weighting (i.e. cost sensitivity) to deal with this problem. 

Class weighting is a popular approach to deal with a class imbalance \cite{akbani2004ApplyingSupportVector}. This way more weight is put on the misclassification of samples from one class (typically the \textit{minority} class) versus the other. 
Adding class weighting to the data fitting term of \eqref{eq:als_step_TKRR} and \eqref{eq:als_step_adapt_tkrr} we obtain,

\begin{align} 
    \begin{split}
    \min_{\tens{W}} \quad & \frac{1}{N} \left[ C^+ \sum^{N^+}_{n^+=1}\left(\left\langle \mat{\Phi}(\vect{x}^+_n),\tens{W}\right\rangle_{\mathrm{F}}-y_n^+\right)^2 \right.\\
    &+ \left. C^- \sum^{N^-}_{n^-=1}\left(\left\langle  \mat{\Phi}(\vect{x}^-_n),\tens{W}\right\rangle_{\mathrm{F}} -y_n^-\right)^2 \right] \\
    &+ \text{regularization}
    \end{split}
    \label{eq:class_weighted_TKRR}
\end{align}
A popular choice for the class weights, and the one used in this paper, is $C^+ = \frac{N}{2N^+}$ and $C^-=\frac{N}{2N^-}$, which reduces to $C^+=C^-=1$ if \mbox{$N^+=N^-=\frac{1}{2}N$}.

Besides using class-weighting, we also undersample the the non-seizure data. This is done until a ratio of 1:10 (seizure to non-seizure data) is obtained. For the patient independent model the same stategy is used an in \cite{vandecasteele2020VisualSeizureAnnotation}. Thus, for every 15 minutes in the first 24 hour period, 10 non-seizure segments are `selected'. These selected samples are then randomly sampled to obtain the desired ratio. This way, we aim to obtain data from all the different brain stages that occur throughout the day.
For the patient-specific or patient-adapted case, the undersampling is done at random per crossvalidation fold.

\subsection{Post-Processing and Model Evaluation} \label{subsec:post_process_eval}
Seizure detection algorithms typically suffer from many false positives, which can create a barrier to the adoption of a wearable \cite{thompson2019SeizureDetectionWatch}. Post-processing can help to reduce these false positive by filtering out `sporadic' positive predictions. 
These strategies are effective because seizures typically are of longer duration than the chosen segment length. A minimum duration of 10 seconds is often considered for automatic seizure detection algorithms. In this paper, we only consider positive detections when at least 8 out of 10 consecutive segments are classified as a seizure \cite{vandecasteele2020VisualSeizureAnnotation}.

Because a seizure is an \textit{event}, it is typically more relevant to consider event-based scores rather than segment-based scores. In this work, we use the any-overlap method to evaluate the event-based performance \cite{shah2021ObjectiveEvaluationMetrics}. The any-overlap method considers a prediction correct if it overlaps with any portion of the true seizure event. The event-based metrics we use are the same as proposed in the SzCORE framework \cite{danSzCORESeizureCommunity}: sensitivity, precision, F1-score and false alarms per day (FA/24hr).

\subsection{Dataset}
For the seizure detection experiments, we use the SeizIT1 behind-the-ear EEG dataset \cite{chatzichristos2023SeizeIT1}. The SeizeIT1 project was approved by the UZ Leuven ethics committee and informed consent was obtained from every participant \cite{vandecasteele2020VisualSeizureAnnotation}. This committee also approved the use of the dataset for the current study (S67350 - amendment 3). Furthermore, approval of the current study was given by the TU Delft Human Research Ethics Committee (HREC) (application ID: 5759). 

The SeizeIT1 data was acquired in the hospital during presurgical evaluation, where patients were monitored with vEEG for multiple days. The data includes behind-the-ear EEG (two electrodes per side), 10-20 full scalp EEG and single-lead ECG. The median duration of the seizures was 50 seconds, with seizures varying from 11 to 695 seconds in length. The seizures consisted mainly of Focal Impared Awareness (FIA) seizures (89\%), with most of the seizures originating from the (fronto-)temporal lobe (91\%). 

Each EEG file is accompanied by two different annotation files. One file (`a1') contains the annotations from the neurologist, which did not always contain an end time for each seizure. The neurologist also noted whether the seizure was visible on the behind-the-ear channels and (if visible) from which hemisphere the seizure originated. The other file (`a2') contains annotations adapted by an engineer \cite{vandecasteele2020VisualSeizureAnnotation} where each seizure has the same length (10 s). Because limiting the seizure duration to only 10 seconds increases the data imbalance even further, we combine the neurologists' annotations with those of the engineer. Thus, in this `combined annotation', each seizure has an end time and a duration of at least 10 seconds. 

For the seizure detection algorithm, we use a bipolar EEG montage that includes a cross-head channel (between the `top' behind-the-ear (bhe) electrodes on each side) and a `lateral' channel. The lateral channel is the difference between the two bhe electrodes on one side of the head. This side is different per patient and depends on the lateralization of the patient's seizures.

\section{Experiments and Results} \label{sec:experiments_results}

The code to obtain the experiments and results presented below can be found on Github\footnote{\url{https://github.com/sderooij/Adapt-TKRR-experiments}}. For the event-based scores, we used the \texttt{timescoring} library from the SzCORE framework with its default settings \cite{danSzCORESeizureCommunity}.

\subsection{Feature Map} \label{subsec:feat_map_hyper_param_tune}
As in \cite{wesel2021LargeScaleLearningFourier, derooij2024EfficientPatientFineTuned} we approximate the RBF kernel with sinusoidal basis functions \cite{solin2020HilbertSpaceMethods}. This means the local feature maps of \eqref{eq:rank1_features} are defined as, 
\begin{equation}
\small{\left(\vect{\phi}^{(d)}\left(x^{(d)}\right)\right)_{i_d}=\frac{1}{\sqrt{U_d}} p\left(\frac{\pi i_d}{2 U_d}\right) \sin \left(\frac{\pi i_d\left(x^{(d)}+U_d\right)}{2 U_d}\right)}, \label{eq:feature_map}
\end{equation}
for $i_d=1, \ldots, {M}_d $, with input data, $x^{(d)} \in [-U_d, U_d]$, and where $p(\cdot)$ is the spectral density of the RBF kernel, \mbox{$p(z) = \sqrt{2\pi \sigma^2} \exp{(-2\pi^2 \sigma^2 z^2)}$} \cite{rasmussen2005GaussianProcessesMachine}, with $\sigma$ as the lengthscale of the RBF kernel.

To simplify things we choose $\forall d, \ M_d = M$ and $ U_d=U$. This leads to three hyperparameters for the feature map, $U$, $M$ and $\sigma$.
How well the feature map approximates the RBF kernel for a given lengthscale depends on the number of basis functions $M$. Generally, for larger $\sigma$, more basis functions are needed because the approximation becomes less accurate close to the boundary. 
Another way to deal with these boundary effects is to increase $U$, while keeping the scaling of the input data the same. This ensures that the input data lies further away from the boundary of the `approximation' interval. 

Below we describe a practical heuristic approach to obtain `good' values for $M$ and $U$ given some $\sigma$:
\begin{enumerate}
    \item In case of a large training set, collect a \textit{small} sample of the training dataset\footnote{We used a sample of 100 datapoints with a 1:1 ratio between positive and negative samples.}, and scale the data to a certain interval (e.g. $[-0.5,0.5]$)
    \item Set $\sigma$ and make a grid for possible values of $M$ and $U$.
    \item For each combination of $M$ and $U$, compute the kernel matrix for the feature map, $\mat{K}_{\Phi}(i,j) = \langle \Phi(\vect{x}_i), \Phi(\vect{x}_j) \rangle_F$.
    \item Compare the $\mat{K}_{\Phi}$ matrices with the true RBF kernel matrix, $\norm{\mat{K}_{\Phi}- \mat{K}_{\text{RBF}}}_\text{F} / \norm{\mat{K}_{\text{RBF}}}_\text{F}$.
    \item Select $M$ and $U$ based on the relative error (and desired model size).
\end{enumerate}

It should be noted that the choice for $M$ changes the size and complexity of the model. Therefore, there might be a trade-off between the `accuracy' of the feature map and the desired model size. 

\subsection{An Example with Synthetic Data} 
\label{subsec:synthetic_data}
\begin{figure}[t]
     \centering
     \vspace{-1.5em}
     \begin{subfigure}[b]{0.24\textwidth}
         \centering
         \includegraphics[width=\textwidth]{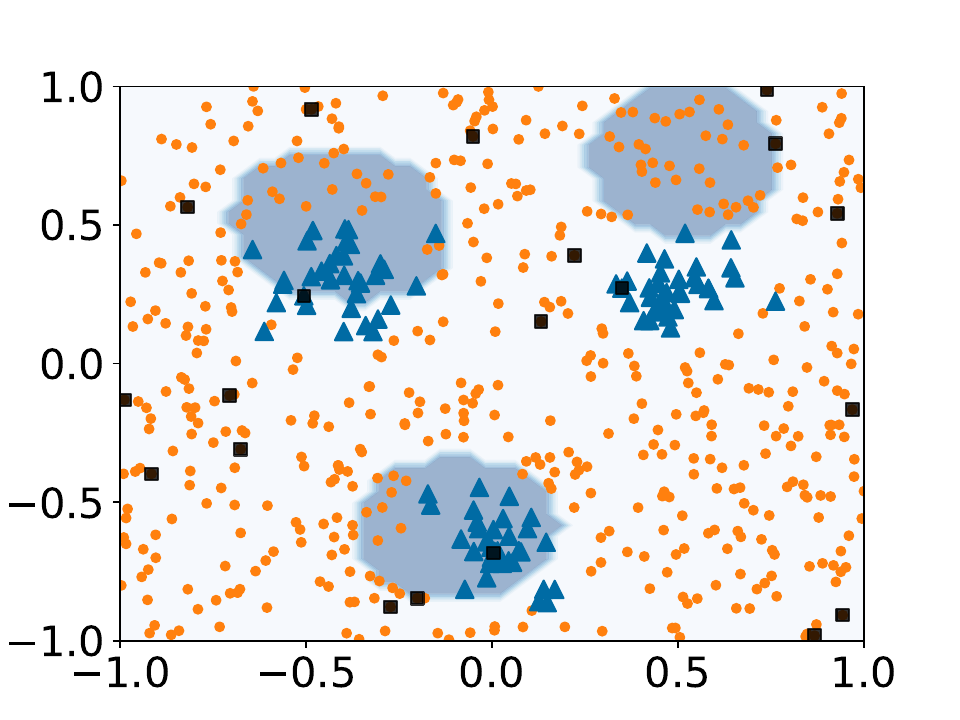}
         \vspace{-1.5em}
         \caption{Source model}
         \label{subfig:source_on_target}
     \end{subfigure}
     \hfill
     \begin{subfigure}[b]{0.24\textwidth}
         \centering
         \includegraphics[width=\textwidth]{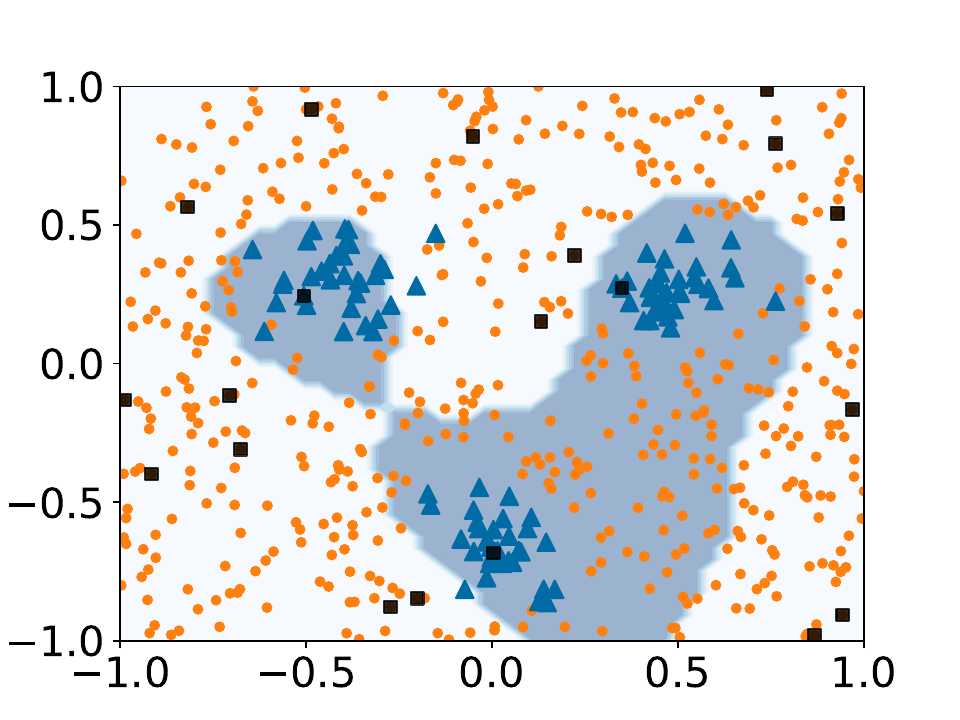}
         \vspace{-1.5em}
         \caption{Target-only model}
         \label{subfig:target_on_target}
     \end{subfigure}
    \begin{subfigure}[b]{0.24\textwidth}
        \centering
        \includegraphics[width=\textwidth]{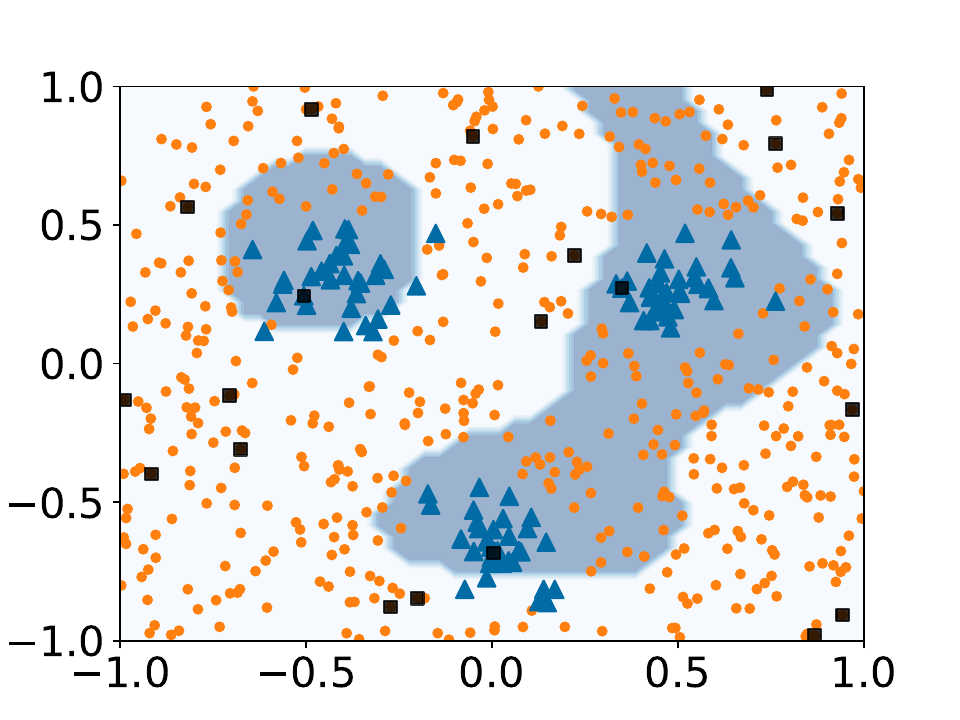}
        \vspace{-1.5em}
        \caption{Adapted model $\mu=10^{-6}$} 
        \label{subfig:adapted_on_target_mu_1e-6}
    \end{subfigure}
         \begin{subfigure}[b]{0.24\textwidth}
         \centering
         \includegraphics[width=\textwidth]{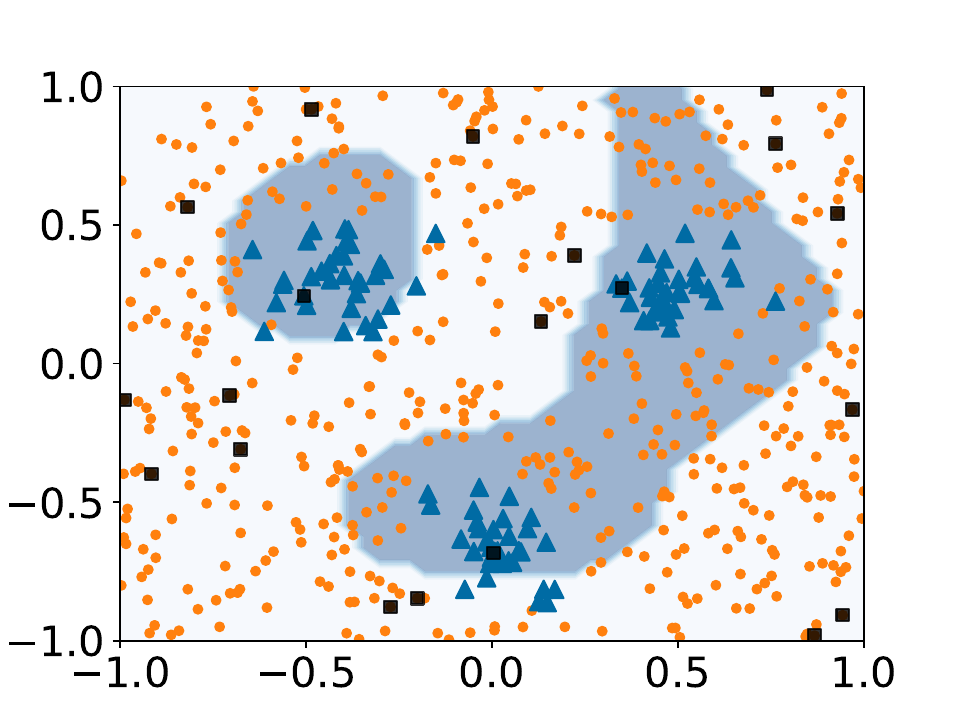}
         \vspace{-1.5em}
         \caption{Adapted model $\mu=10^{-4}$} 
         \label{subfig:adapted_on_target_mu_0.0001}
     \end{subfigure}
     \hfill
         \begin{subfigure}[b]{0.24\textwidth}
         \centering
         \includegraphics[width=\textwidth]{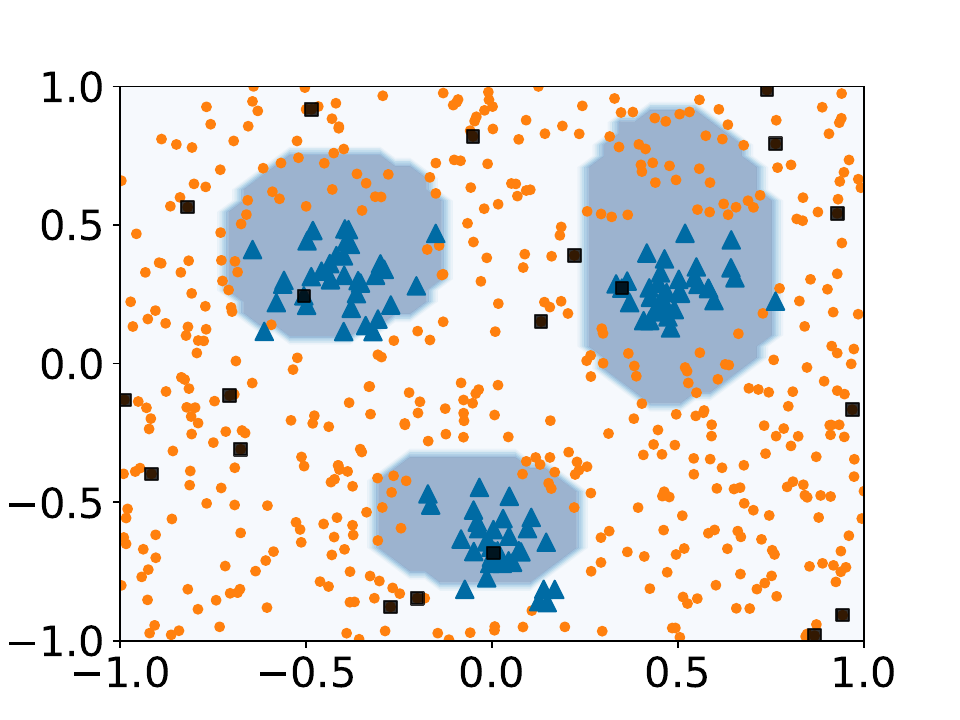}
         \vspace{-1.5em}
         \caption{Adapted model $\mu=0.01$} 
         \label{subfig:adapted_on_target}
     \end{subfigure}
     \hfill
        \begin{subfigure}[b]{0.24\textwidth}
        \centering
        \includegraphics[width=\textwidth]{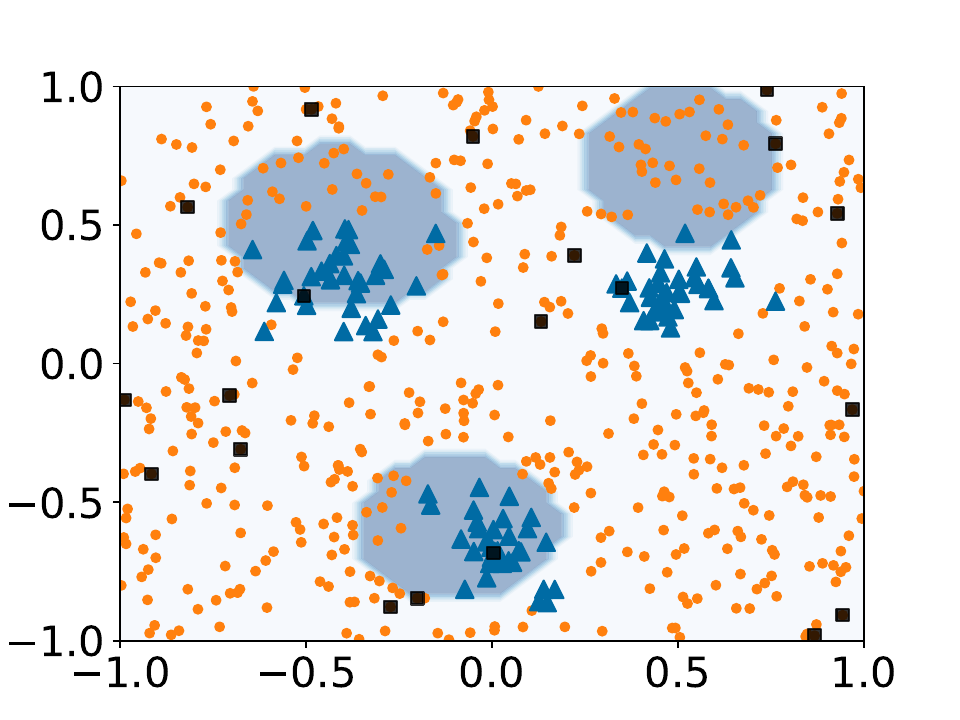}
        \vspace{-1.5em}
        \caption{Adapted model $\mu=1$} 
        \label{subfig:adapted_on_target_mu_1}
    \end{subfigure}
        \caption{Decision boundaries of different TKRR models on the synthetic target data. The orange dots show the negative samples, the blue triangles the positive samples and the samples used for training are highlighted by the black squares. The shaded blue area shows where the classifier labels samples as \textit{positive}.}
        \label{fig:decision_boundaries_synthetic_data}
\end{figure}
To showcase the effectiveness of the Adapt-TKM algorithm, we first apply it to some synthetically generated data. We use the same strategy as in the Adapt-SVM paper \cite{yang2007AdaptingSVMClassifiers}. Thus, we generate the positive data for the source and target dataset from a Gaussian mixture with 3 components, and the negative data is sampled from a uniform distribution outside of the area of the positive data. In the source dataset, the Gaussian components are centered at $(-0.4, 0.5)$, $(0.5, 0.7)$ and $(-0.1, -0.6)$, while in the target dataset, the means are at $(-0.4, 0.3)$, $(0.5, 0.3)$ and $(0, -0.65)$. Each dataset has $500$ negative and $100$ positive samples. 

First, we train a source model using the TKRR classifier with the aforementioned feature map \eqref{eq:feature_map}. Due to the data imbalance, class weighting was used as described in Section \ref{subsec:class_weighting_undersamp}. The feature map parameter $\sigma$ was chosen to coincide with the RBF parameter used in \cite{yang2007AdaptingSVMClassifiers}\footnote{In \cite{yang2007AdaptingSVMClassifiers} $\kappa(x_i,x_j) = e^{-\rho \norm{x_i-x_j}^2}$ with $\rho=5$, so $\sigma = \sqrt{\frac{1}{2\cdot\rho}} = \sqrt{\frac{1}{10}}$ .}.  Then, the other feature map parameters ($M$ and $U$) are determined using the procedure described in the previous section.
In Table \ref{tab:rel_error_feature_map_synthetic_source}, we show the relative error between the true RBF kernel matrix, $K_{\text{RBF}}$ and the one calculated with the feature map, $K_{\mat{\Phi}}$. Based on these values, $M=14$ and $U=1.75$ were selected, as the error for these values was deemed sufficiently low. The rank of the CPD weight tensor was set to $R=4$ and the regularization parameter to $\lambda=0.001$. Figure \ref{subfig:source_on_target} shows the decision boundary of the source model applied to the target dataset. 

\begin{table}[t]
\centering
\renewcommand{\arraystretch}{1.5}
\resizebox{0.485\textwidth}{!}{
\begin{tabular}{|c||c|c|c|c|c|c|}
\hline
\diagbox{$M$}{$U$} & 1 & 1.25 & 1.5 & 1.75 & 2 & 2.25  \\ \hline \hline
10  & $0.42$ & $0.051$ & $1.3 \cdot 10^{-3}$ & $5.6 \cdot 10^{-3}$ & $1.6 \cdot 10^{-2}$ & $3.7 \cdot 10^{-2}$  \\ \hline
11  & $0.42$ & $0.051$ & $8.1 \cdot 10^{-4}$ & $2.2 \cdot 10^{-3}$ & $8.0 \cdot 10^{-3}$ & $2.0 \cdot 10^{-2}$  \\ \hline
12  & $0.42$ & $0.051$ & $7.4 \cdot 10^{-4}$ & $7.5 \cdot 10^{-4}$ & $3.7 \cdot 10^{-3}$ & $1.0 \cdot 10^{-2}$ \\ \hline
13  & $0.42$ & $0.051$ & $7.3 \cdot 10^{-4}$ & $2.5 \cdot 10^{-4}$ & $1.5 \cdot 10^{-3}$ & $5.2 \cdot 10^{-3}$  \\ \hline
14  & $0.42$ & $0.051$ & $7.3 \cdot 10^{-4}$ & $8.0 \cdot 10^{-5}$ & $6.0 \cdot 10^{-4}$ & $2.6 \cdot 10^{-3}$  \\ \hline
15  & $0.42$ & $0.051$ & $7.2 \cdot 10^{-4}$ & $2.3 \cdot 10^{-5}$ & $2.3 \cdot 10^{-4}$ & $1.2 \cdot 10^{-3}$  \\ \hline
16  & $0.42$ & $0.051$ & $7.2 \cdot 10^{-4}$ & $5.8 \cdot 10^{-6}$ & $8.5 \cdot 10^{-5}$ & $5.1 \cdot 10^{-4}$  \\ \hline
17  & $0.42$ & $0.051$ & $7.2 \cdot 10^{-4}$ & $1.7 \cdot 10^{-6}$ & $2.8\cdot 10^{-5}$ & $2.1 \cdot 10^{-4}$  \\ \hline
18  & $0.42$ & $0.051$ & $7.2 \cdot 10^{-4}$ & $1.2 \cdot 10^{-6}$ & $8.4 \cdot 10^{-6}$ & $8.8 \cdot 10^{-5}$ \\ \hline
19  & $0.42$ & $0.051$ & $7.2 \cdot 10^{-4}$ & $1.0 \cdot 10^{-6}$ & $2.6 \cdot 10^{-6}$ & $3.3 \cdot 10^{-5}$  \\ \hline
20  & $0.42$ & $0.051$ & $7.2 \cdot 10^{-4}$ & $9.9 \cdot 10^{-7}$ & $7.4 \cdot 10^{-7}$ & $1.2 \cdot 10^{-5}$  \\ \hline
 \end{tabular}}
\caption{Relative error, $\norm{\mat{K}_{\Phi}- \mat{K}_{\text{RBF}}}_\text{F} / \norm{\mat{K}_{\text{RBF}}}_\text{F}$, at different values for $M$ and $U$ on the synthetic source dataset.}
\label{tab:rel_error_feature_map_synthetic_source}
\end{table}

The same model parameters were used to train a target-only TKRR model (Figure \ref{subfig:target_on_target}) using a small subset of samples from the target dataset: $17$ negative and $3$ positive samples. This same subset was used to train the Adapt-TKRR model for varying values of $\mu$ (Figures \ref{subfig:adapted_on_target_mu_1e-6}-\ref{subfig:adapted_on_target_mu_1}).

From Figure \ref{fig:decision_boundaries_synthetic_data}, it is clear that the adapted model becomes more similar to the source model for high $\mu$ and more similar to the target model for low $\mu$. The best performance is obtained at $\mu=0.01$. In this case, the F1-score is $0.67$, which is better than both the source ($0.48$) and target-only model's ($0.61$) F1-scores.

\subsection{Performance of Adapt-TKRR on Seizure Detection}
\begin{table}[t]
\centering
\renewcommand{\arraystretch}{1.5}
\resizebox{0.489\textwidth}{!}{
\begin{tabular}{|c||c|c|c|c|c|c|}
\hline
Model & Sensitivity & Precision & F1-score  & FA/24hr \\ \hline \hline
SVM-PI & 0.620 (0.22) & 0.548 (0.42) & 0.458 (0.31) & 9.06 (14) \\ \hline
TKRR-PI & 0.630 (0.18) & 0.575 (0.44) & 0.486 (0.33) & 12.0 (23) \\ \hline
SVM-PS & 0.623 (0.12) & 0.508 (0.31) & 0.443 (0.21) & 16.6 (22) \\ \hline
TKRR-PS & 0.625 (0.14) & 0.414 (0.35) & 0.347 (0.25) & 26.2 (30) \\ \hline
SVM-PA & 0.621 (0.15)& 0.555 (0.33) & 0.484 (0.22) & 6.5 (7.1) \\ \hline
TKRR-PA & 0.632 (0.17) & 0.601 (0.37) & 0.523 (0.28) & 5.19 (7.3) \\ \hline
  \end{tabular}}
\caption{Average performance across the patients, standard deviation between brackets.}
\label{tab:average_results_over_patients}
\end{table}
\begin{figure}[t]
    \centering
    \includegraphics[width=0.49\textwidth]{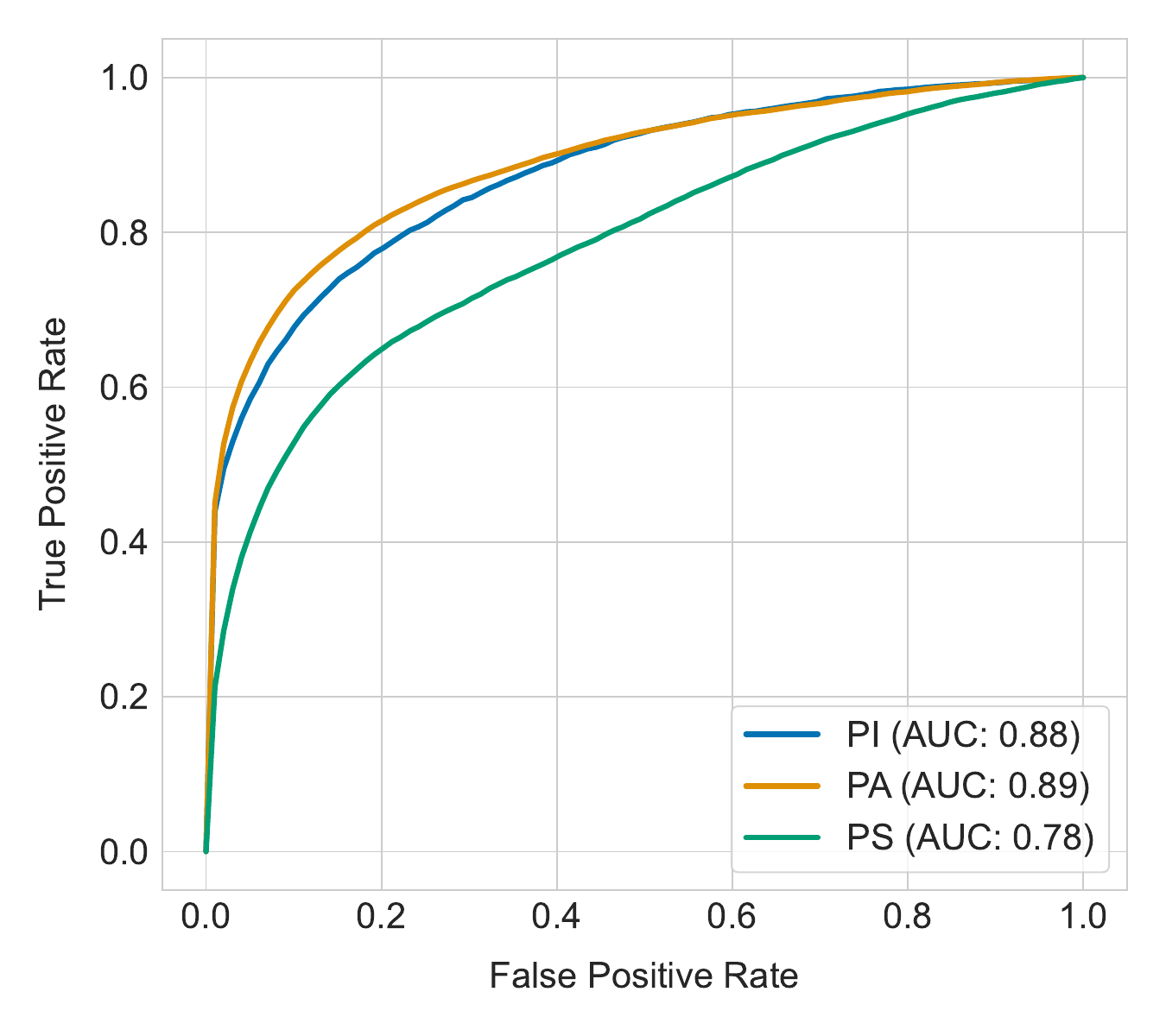}
    \caption{ROC curves of the TKRR models}
    \label{fig:ROC_TKRR_models}
\end{figure}

\begin{figure}[t]
    \centering
    \includegraphics[width=0.49\textwidth]{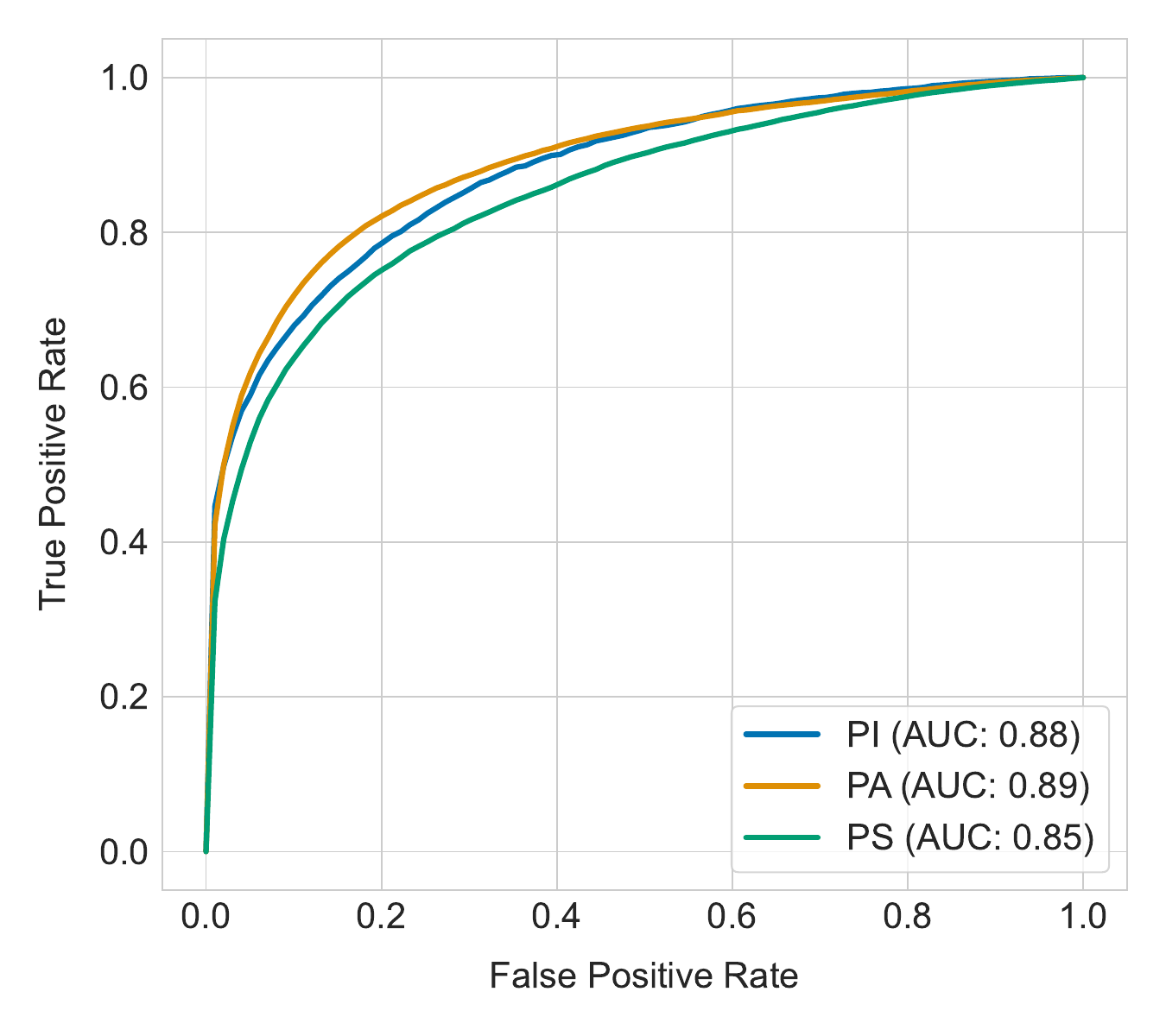}
    \caption{ROC curves of the SVM models}
    \label{fig:ROC_SVM_models}
\end{figure}
\begin{figure*}[t]
    \centering
    \includegraphics[width=\textwidth]{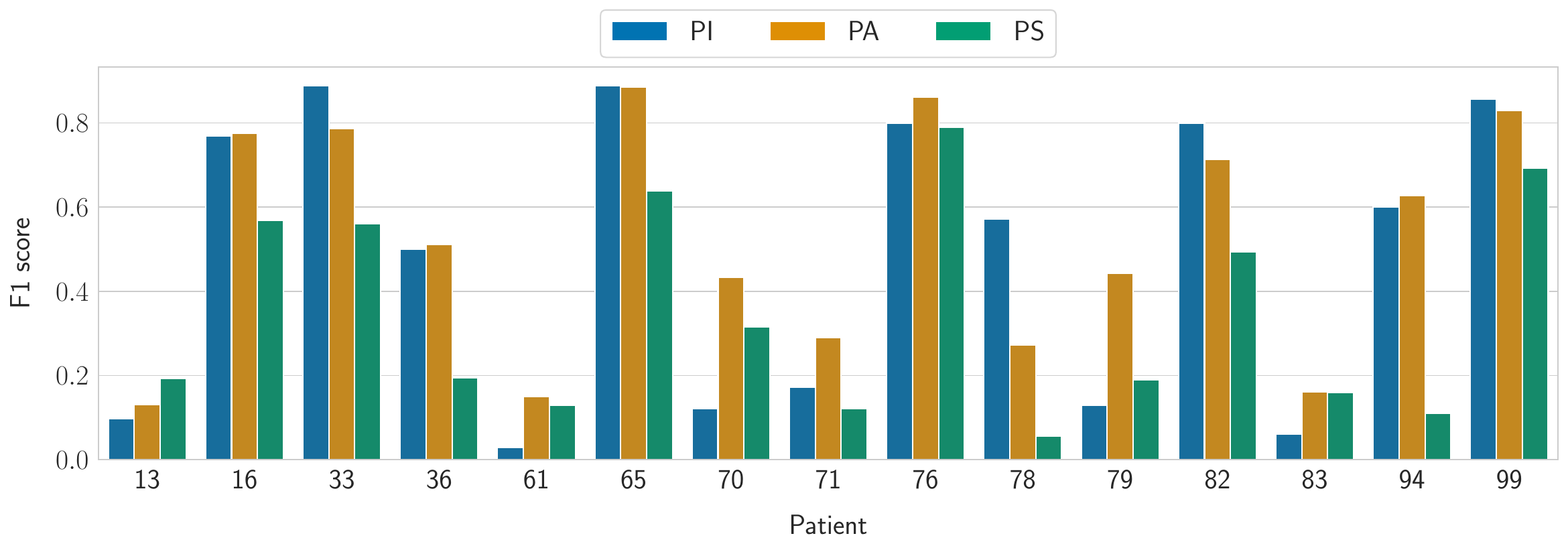}
    \caption{F1-scores per patient for the TKRR models.}
    \label{fig:barplot_f1_per_patient}
\end{figure*}
\begin{figure*}[t]
    \centering
\includegraphics[width=\textwidth]{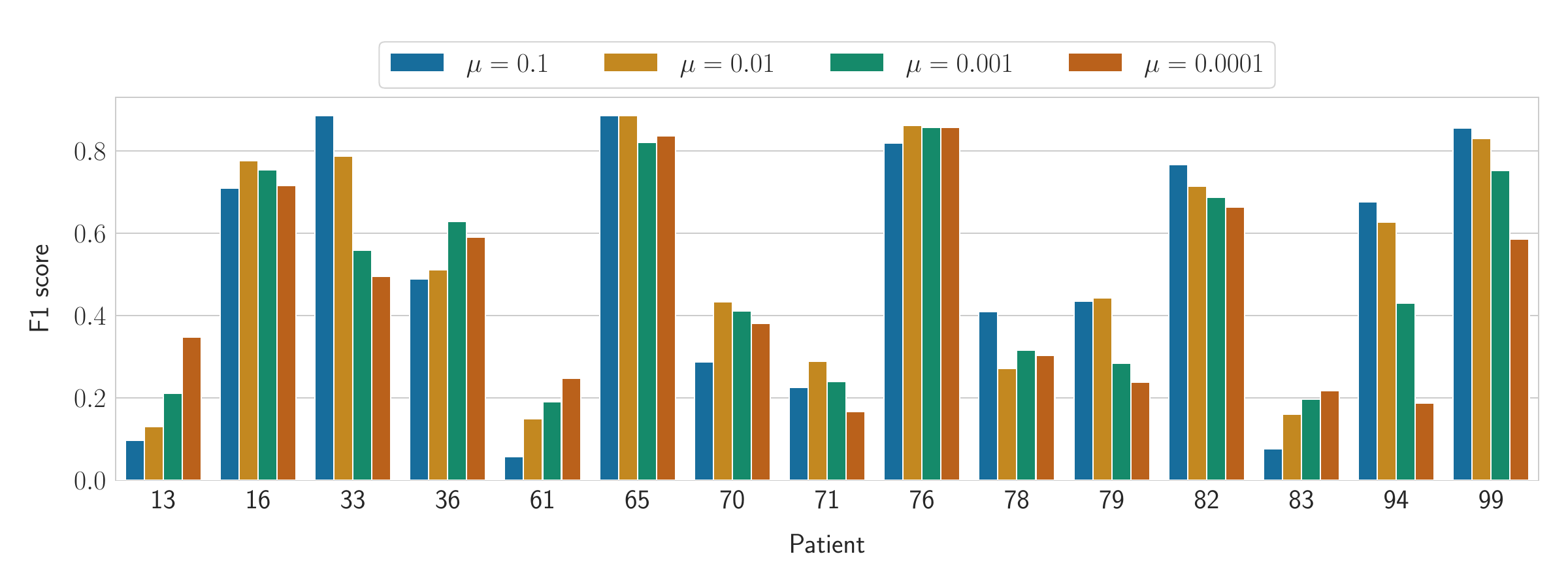}
    \caption{F1-scores per patient for different values of $\mu$ in the Adapt-TKRR model.}
\label{fig:barplot_diff_mu_per_patient}
\end{figure*}
In this section, we test the effectiveness of the Adapt-TKRR model in personalizing seizure detection models. Thus, we first implement a patient-independent (i.e. \textit{source}) model (TKRR-PI), which is then \textit{adapted} using a small amount of patient-specific (or \textit{target}) data. The performance of these patient-\textit{adapted} models (TKRR-PA) is also compared to patient-specific models (TKRR-PS), which are trained only on the patient-specific data. As a reference, we compare the performance of the tensor kernel machines to the performance of the SVM and Adapt-SVM classifiers. 

The patient-independent (PI) models are validated using a leave-one-patient-out cross-validation procedure. Furthermore, we only use the seizures visible on the behind-the-ear channels to train the PI models, as we found that this led to the best results. For testing, we did not adhere to these restrictions, so the models were tested on all the seizures. Hyperparameter tuning for the PI models was done using five-fold cross-validation on the training sets, maximizing the area under the ROC curve (AUROC). 

The patient-adapted (PA) and patient-specific (PS) models are validated using a leave-one-seizure-\textit{in} cross-validation procedure. The splits of the cross-validation folds are set in the middle between two seizures. Thus, every fold contains one seizure with varying
amounts of non-seizure data. As stated in Section \ref{subsec:class_weighting_undersamp}, this non-seizure data is randomly sampled to obtain the desired ratio for the training set. Furthermore, we only consider patients who experienced five or more seizures in order to have sufficient data for testing.

For the PS and PA models, the training datasets were too small to obtain `good' hyperparameters through cross-validation on the training data. Thus, we used the same hyperparameters as for the corresponding PI model. Because of the difference in regularization, the regularization parameter of the adaptive models could not be directly inherited from the PI model. Thus, we set them to a fixed value:  $\mu = 0.01$ (for Adapt-TKRR, same as the optimum for the synthetic data \ref{subsec:synthetic_data}) and $C=10$ (for Adapt-SVM, same as in \cite{yang2007AdaptingSVMClassifiers}). 
In the next section (\ref{sec:influence_mu_adapt_tkrr}), we will discuss the influence of the $\mu$ hyperparameter for the TKRR-PA models more in depth.

Table \ref{tab:average_results_over_patients} shows the average performance across the patients for the different models. The scores presented are the event-based metrics (so after post-processing as described in Section \ref{subsec:post_process_eval}). To better compare the performance of the different models, we tuned the thresholds to obtain similar sensitivity levels. 
Thus, we can see that the PA models have a lower false positive rate at comparable sensitivity to the PI models. On the other hand, the TKRR-PS model has the worst performance across all metrics. 

This result is supported by the difference in ROC curves, which are plotted in Figure \ref{fig:ROC_TKRR_models} for the TKRR models and in Figure \ref{fig:ROC_SVM_models} for the SVM models. From these ROC curves, it is also clear that when it comes to the performance of the PI and PA models, the (Adapt-)TKRR and (Adapt-)SVM classifiers lead to fairly similar results. The TKRR-PS model, however, performs much worse than its SVM counterpart. This is likely due to the inheritance of the model parameters from the PI model, which for the TKRR classifier also means inheriting its size. 

So far, the results presented have been averaged over the patients. It might, however, be more interesting to see the patient-by-patient results. To this end, we compute the (event-based) F1-scores per patient for the TKRR models. These are shown in Figure \ref{fig:barplot_f1_per_patient}. Here, we can see that the patient-adapted model performs the best for 9 out of 15 patients. There are, however, also patients for which the PA model decreases the performance of the PI model (patients 33, 65, 78, 82 and 99). These seem to be patients where the PI model performs relatively well (F1-score $>0.6$), and the PS model performs significantly worse, suggesting that adaptation may have a negative effect here. 

\subsection{Influence of the Regularization Parameter} \label{sec:influence_mu_adapt_tkrr}
As previously seen on the synthetic data in Section \ref{subsec:synthetic_data}, the choice for the regularization parameter, $\mu$, significantly influences the performance of the Adapt-TKRR model. To investigate the influence of these parameters on the performance, we compare the F1-scores for each patient for different $\mu$ values in Figure \ref{fig:barplot_diff_mu_per_patient}. 
From this Figure, it is clear that some patients (33, 65, 78, 79, 82, 94 and 99) benefit from high $\mu$ values, while for others, the Adapt-TKRR performs best when $\mu$ is more `moderate' (16, 36, 70, 71 and 76) or low (13, 61 and 83).  

Typically, the patients for whom a high $\mu$ value is optimal are the ones where the performance of the PI model is already `good'. Whereas the ones that benefit from lower $\mu$ values are typically those for whom the PS model performs (relatively) better. If we choose the optimal value for $\mu$ for each patient (based on Figure \ref{fig:barplot_diff_mu_per_patient}), we can reach performance that is better than (or at least on par with) the PI and PS models for all but two patients. For these two patients (78 82), the $\mu$ value needs to be even higher than $0.1$ to reach the same performance as the PI model, which essentially means that no transfer occurs and the model stays the `same' (such as the case for $\mu=1$ on the synthetic data, Figure \ref{subfig:adapted_on_target_mu_1}). 

Besides the patient-by-patient results, we also look at the overall performance of the TKRR-PA models for different $\mu$ values. The event-based scores of these models are found in Table \ref{tab:mean_results_different_mu}. In this table, we also present the performance in the case where the `optimal' $\mu$ is chosen for each patient (based on Figure \ref{fig:barplot_diff_mu_per_patient}). As expected, this `optimal' model has the best performance among all metrics. Interestingly, the next best performing model is the one where $\mu=0.01$, which corresponds with the synthetic data experiment (and was therefore chosen for the TKRR-PA model). This suggests that selecting $\mu=0.01$ for the Adapt-TKRR model provides a reasonable trade-off between source and target if no prior information is available. 

\begin{table}[t]
\centering
\renewcommand{\arraystretch}{1.5}
\resizebox{0.489\textwidth}{!}{
\begin{tabular}{|c||c|c|c|c|c|c|}
\hline
$\mu$ & Sensitivity & Precision & F1-score  & FA/24hr \\ \hline \hline
$1\times 10^{-4}$ & 0.628 (0.15) & 0.508 (0.33) & 0.456 (0.24) & 11.1 (12) \\ \hline
$1 \times 10^{-3}$ & 0.636 (0.15) & 0.552 (0.34) & 0.490 (0.25) & 5.54 (5.5) \\ \hline
$0.01$ & 0.632 (0.17) & 0.601 (0.37) & 0.523 (0.28) & 5.19 (7.3)  \\ \hline
$0.1$ & 0.635 (0.19) & 0.608 (0.41) & 0.512 (0.31) & 8.04 (13) \\ \hline
`Optimal' & 0.648 (0.16) & 0.671 (0.35) & 0.582 (0.25) & 4.19 (7.1) \\ \hline
  \end{tabular}}
\caption{Average performance (and standard deviation) across the patients of the TKRR-PA model for different $\mu$ values. `Optimal' is the case when using the optimal value for each patient based on the F1-score.}
\label{tab:mean_results_different_mu}
\end{table}

\subsection{Initialization of the Adapt-TKM Model}

\begin{figure}[t]
    \centering
    \includegraphics[width=0.49\textwidth]{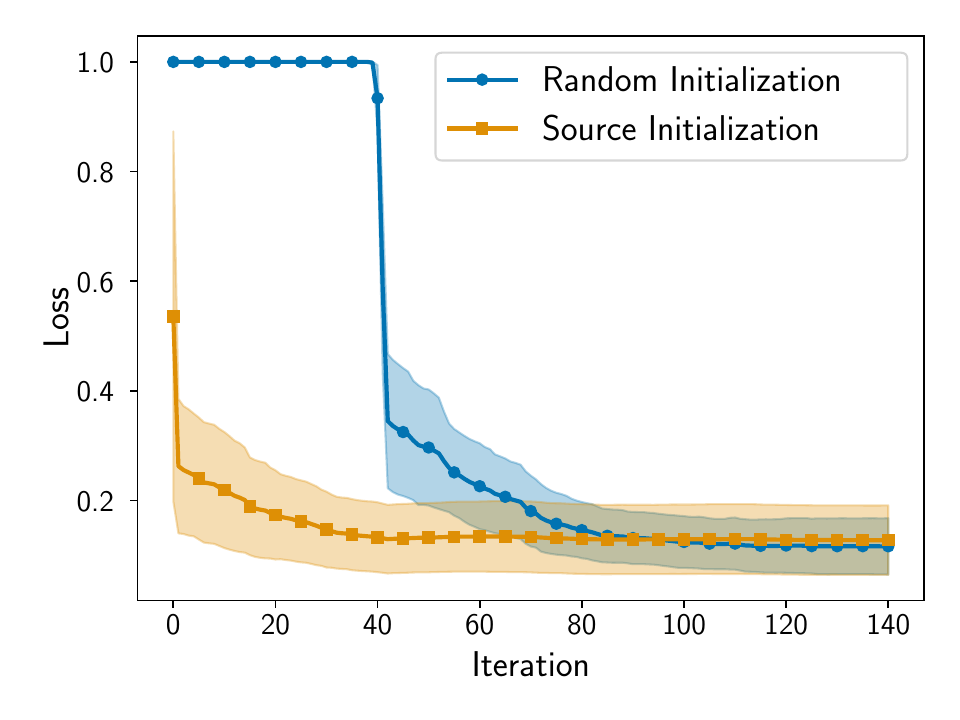}
    \caption{Mean convergence of the Adapt-TKRR models (PA models) for initialization with the source model (PI model) and random initialization. The y-axis shows the loss (i.e. the data fitting term of \eqref{eq:class_weighted_TKRR}), and the x-axis shows the iteration count. The shaded areas show the spread of $\pm 1$  standard deviation. Dots are placed every 5 iterations.}
    \label{fig:convergence_adapt_tkrr}
\end{figure}

In this section, we investigate whether initialization of the Adapt-TKRR model with the source model weights improves convergence. In our previous work \cite{derooij2024EfficientPatientFineTuned}, we found that initialising a patient-specific TKRR model with the PI weights improved the convergence speed significantly (`patient-\textit{finetuning}'). To test whether this also holds for the Adapt-TKRR models, we plot the normalized training loss of the TKRR-PA models against the number of iterations (Figure \ref{fig:convergence_adapt_tkrr}). The loss is computed by calculating the data fitting term of \eqref{eq:class_weighted_TKRR}, i.e. $\ell = \frac{1}{N} [ C^+ \sum^{N^+}_{n^+=1}\left(\left\langle \mat{\Phi}(\vect{x}^+_n),\tens{W}\right\rangle_{\mathrm{F}}-y_n^+\right)^2 +  C^- \sum^{N^-}_{n^-=1}\left(\left\langle  \mat{\Phi}(\vect{x}^-_n),\tens{W}\right\rangle_{\mathrm{F}} -y_n^-\right)^2 ]$. 

The models using random initialization start to converge after around 42 iterations; this equals exactly one full sweep over the factor matrices of the CPD weights ($D=42$). The optimum for the randomly initialized models is not reached until about 110 iterations are done. The models initialized with the source weights, on the other hand, reach their optimum already after about one sweep (42 iterations). Random initialization does seem to lead to a better optimum than initialization with source weights, though the difference is small: $0.011$. 

\subsection{Model Efficiency}
To analyze the efficiency of the models, we compare the total number of parameters and the inference speed for each model. 
The number of model parameters for the SVM models are equal to the size of support vectors ($N_{sv} D$) plus the dual coefficients ($N_{sv}$). The amount of model parameters for the TKRR models is simply determined by the size of the CPD weight tensor ($DMR$).

The second column of Table \ref{tab:model_parameters_inference_time} presents the average number of parameters for each model.  It should be noted that because the model hyperparameters for the TKRR models were kept the same for the PA and PS models as the corresponding PI model, the amount of model hyperparameters is the same across the models. From these values, it is clear that the TKRR models have about 100 times fewer model parameters than the SVM-PI model. Only the SVM-PS models have slightly fewer hyperparameters, though it is still in the same order of magnitude ($10^4$) as the TKRR models. 

The inference speeds were determined by randomly selecting $1000$ samples (same for each model) and computing the average time it takes the model to determine the label for a sample. The computations were performed on the same hardware: a laptop computer with a quad-core Intel i7-1185G7 (3.0 GHz) processor and 32GB of RAM. The third column of Table \ref{tab:model_parameters_inference_time} presents the average inference speed per model. Again, we see that the SVM-PI and SVM-PA models are the least efficient. They are about 100 times slower than the TKRR and the SVM-PS models.

\begin{table}[t]
\centering
\renewcommand{\arraystretch}{1.5}
\begin{tabular}{|c||c|c|}
\hline
Model & No. Parameters & Inference Time (s) \\ \hline \hline
SVM-PI & $1.363 \cdot 10^6$ & $3.3 \cdot 10^{-3}$ \\ \hline
TKRR-PI & $1.988 \cdot 10^4$ & $3.9 \cdot 10^{-5}$ \\ \hline
SVM-PS & $1.140 \cdot 10^4$ & $3.3 \cdot 10^{-5}$ \\ \hline
TKRR-PS & $1.988 \cdot 10^4$ & $3.9 \cdot 10^{-5}$ \\ \hline
SVM-PA & $1.369 \cdot 10^6$ & $5.2 \cdot 10^{-3}$ \\ \hline
TKRR-PA & $1.988 \cdot 10^4$ & $3.9 \cdot 10^{-5}$ \\ \hline
\end{tabular}
\caption{Average number of parameters and inference time (in seconds) per model.}
\label{tab:model_parameters_inference_time}
\end{table}

\section{Discussion and Future Work} \label{sec:discussion}
From the results presented in the previous section, it is clear that the adaptive tensor kernel machine can be an effective and efficient transfer learning model. It is able to improve the seizure detection performance (for some patients) by personalizing a patient-independent model. Furthermore, it can do this more efficiently than the Adapt-SVM model and without increasing the model size. Besides the efficiency gain, there is also another advantage to the Adapt-TKM model over the Adapt-SVM. The Adapt-SVM uses the support vectors of the source model, these support vectors are the training data of the source model. In the seizure detection case, this may raise privacy concerns due to `saving' data from other patients. The Adapt-TKM model does not `save' any source data and is thus more `privacy-preserving' \cite{zhao2023SourceFreeDomainAdaptation}.

While we have seen that the Adapt-TKM can be effective in personalizing seizure detection models, there are still limitations to its effectiveness. Most notable is that seizure detection performance did not improve for all patients. Especially for patients who already had good performance from the PI models the adaptation could have a negative effect.
Part of this negative effect is because choosing the regularization of the Adapt-TKM model (or the rate of transfer) is not straightforward. While the value we chose ($\mu =0.01$) seemed to have decent performance overall, this did not hold for all cases (Figure \ref{fig:barplot_diff_mu_per_patient}).  Thus, finding an effective strategy to choose this parameter remains a topic of future research. 

Another limitation of the Adapt-TKM model, or actually tensor kernel machines in general, is that it introduces extra hyperparameters compared to `standard' kernel machines. For the extra hyperparameters of the feature map, $M$ and $U$, we propose a heuristic approach in Section \ref{subsec:feat_map_hyper_param_tune}. However, there is no straightforward way to select the rank, $R$, of the CPD weight tensor, and it was done via a grid search on the PI data. Inheriting this hyperparameter from the PI model may have been suboptimal for the PA and PS models and could have led to larger models than necessary.
A more informed way to choose (or update) the rank of the CPD weight tensor, may improve performance and reduce the need for an exhaustive grid search.

\section{Conclusion} \label{sec:conclusion}
In this work, we have presented an efficient transfer learning method using tensor kernel machines. This method, the adaptive tensor kernel machine (Adapt-TKM), can adapt a previously learned source model to new data. It does so by transferring `knowledge' from the source weights through regularization while learning the adapted model weights. The advantage of this Adapt-TKM model over the conventional adaptive SVM is that it is able to learn a compact non-linear model in the primal, using low-rank tensor networks to deal with the high-dimensionality of the feature space.   
This way the Adapt-TKM does not increase in size compared to the source, as opposed to the kernelized adaptive SVM.

We showcased the effectiveness of the method by applying it to seizure detection for wearable behind-the-ear EEG. The Adapt-TKM model efficiently personalizes a patient-independent model with minimal patient-specific data. This patient-adapted model outperformed both the patient-independent and patient-specific models. Moreover, the Adapt-TKM model was about a hundred times more efficient than its SVM counterpart, both in terms of the number of parameters and its inference speed.

\section*{Acknowledgements}
We would like to thank both Kaat Vandecasteele and Miguel Bhagubai for answering our questions related to the seizure detection pipeline for the SeizeIT1 dataset. 
We would also like to thank Javier Macea for providing some clinical insight into the dataset.

\bibliographystyle{IEEEtran}
\bibliography{Bibliography.bib}

\end{document}